\documentclass[sigconf,screen]{acmart}

\usepackage{multirow, tabularx, dblfloatfix}
\newcolumntype{C}{>{\centering\arraybackslash}X}

\AtBeginDocument{%
  }

\setcopyright{acmcopyright}
\copyrightyear{2023}
\acmYear{2023}
\acmDOI{XXXX/XXXX.XXXX}

\acmConference[ACM MM '23]{ACM MM '23: ACM International Conference on Multimedia}{October 29, 2023 – November 3, 2023}{Ottawa, Canada}
\acmPrice{15.00}
\acmISBN{978-1-4503-XXXX-X/18/06}

\begin{document}

\title{NNVISR: Bring Neural Network Video Interpolation and Super Resolution into Video Processing Framework}

\author{Yuan Tong}
\author{Mengshun Hu}
\author{Zheng Wang}
\authornote{Corresponding author}
\affiliation{%
  \institution{National Engineering Research Center for Multimedia Software,
  Hubei Key Laboratory of Multimedia and Network Communication Engineering,
  School of Computer Science, Wuhan University, China}
  \country{}
}

\renewcommand{\shortauthors}{Tong et al.}

\begin{abstract}
We present NNVISR - an open-source filter plugin for the
VapourSynth\footnote{https://github.com/vapoursynth/vapoursynth}
video processing framework, which facilitates the application of neural networks for various kinds of video enhancing tasks, including denoising, super resolution, interpolation, and spatio-temporal super-resolution. NNVISR fills the gap between video enhancement neural networks and video processing pipelines, by accepting any network that enhances a group of frames, and handling all other network agnostic details during video processing. NNVISR is publicly released at https://github.com/tongyuantongyu/vs-NNVISR.
\end{abstract}




\maketitle

\section{Introduction}

In recent years, deep learning has brought tremendous advancements in
fundamental computer vision tasks, such as
image classification \cite{10.1145/3065386,He_2016_CVPR,pmlr-v97-tan19a},
object detection \cite{NIPS2015_14bfa6bb,Redmon_2016_CVPR, 10.1007/978-3-030-58452-8_13}
and image super resolution \cite{7115171,Ledig_2017_CVPR,Wang_2018_ECCV_Workshops}.
These advancements also shed light to video enhancing tasks, such as
super resolution \cite{Caballero_2017_CVPR,Tao_2017_ICCV, Wang_2019_CVPR_Workshops,Tian_2020_CVPR,Chan_2021_CVPR},
interpolation \cite{Jiang_2018_CVPR,Bao_2019_CVPR,NEURIPS2019_d045c59a,8840983,Choi_Kim_Han_Xu_Lee_2020, huang2022rife},
and spatio-temporal super resolution \cite{10.1145/3503161.3547874,Hu_2022_CVPR,Xue2019, Xiang_2020_CVPR,Xu_2021_CVPR,Haris_2020_CVPR,Dutta_2021_CVPR}.

However, different from the fundamental tasks which deal with
single images, video enhancing tasks deal with videos,
which are indefinitely long sequences of images
(which is usually called frame), and can be extraordinarily large
when in uncompressed form.
Such intrinsic characteristics of videos decided that it is
unfeasible to process videos like images which load the whole input into computational memory. 
In fact, uncompressed videos are so large that they can quickly
fill up even storage devices, let along memory.
To solve this problem, many video enhancement methods
\cite{Caballero_2017_CVPR, Tao_2017_ICCV, Xue2019, Wang_2019_CVPR_Workshops, 10.1145/3503161.3547874, Hu_2022_CVPR, Choi_Kim_Han_Xu_Lee_2020, Haris_2020_CVPR, Bao_2019_CVPR, Xiang_2020_CVPR, Xu_2021_CVPR}
actually works on a definite and relatively small number of
consecutive frames, thus limiting the amount of data to be processed at a time, while still being able to explore the extra temporal relation in video.

Such change to the task allowed applying deep learning method
to video enhancing tasks, but also left the ultimate task of
enhancing videos in an half-done state, compared to image tasks
where task and network inputs and outputs are aligned and can
be directly integrated into current applications. Most video
enhancement network implementations expect input to be a tuple
of images, so to utilize them one have to first convert video
to image file sequence and separate them into tuples, then run the network inference, and finally merge output tuples into sequence and convert to video, which is both inefficient and resource-consuming. Some network implementations do try to support process video by using Python libraries like OpenCV, but is barely usable and far from the requirement of serious video processing, let along performance. Moreover, by processing frame tuples instead of the whole video, video enhancement networks avoided a non-trivial problem of detecting scene changes, across which exploring temporal relation usually brings more noise than information and should generally be avoided.

In developing a practical spatio-temporal video super resolution tool based on our previous work CycMuNet~\cite{Hu_2022_CVPR} and YOGO~\cite{10.1145/3503161.3547874},
we found that most of the work is not specific to our
network, and even not specific to the spatio-temporal video
super resolution task. Instead, any video enhancement network
that produces output frames based on a consecutive tuple of
input frames can fit into our design. This motivated us to develop
NNVISR, an open-source tool, brings all kinds of video enhancement networks into the video processing framework.

NNVISR is publicly released at https://github.com/tongyuantongyu/vs-NNVISR under the
3-Clause BSD License. The repository contains all the source code
and detailed documentation including installation, compilation and
network intergration guide. NNVISR also provides binary release
on Anaconda, which can be easily installed along with all necessary
dependencies using the conda package manager. NNVISR is designed
as a filter plugin for the open source
VapourSynth
video processing framework. VapourSynth uses Python grammar to
declare processing pipeline, and has mature support for
loading video, traditional video processing and piping output
to video encoders like FFmpeg and x264. NNVISR uses NVIDIA's
TensorRT\footnote{https://developer.nvidia.com/tensorrt} framework
to provide high performance inference of deep learning network.
Therefore, NNVISR is suitable for both quick demo and large-scale
deployment. To the best of our knowledge, NNVISR is the first
video processing tool to support inter-frame video
enhancement using arbitrary neural network.

\section{Related Work}

\subsection{Video Super Resolution}

Video Super Resolution aims to convert video frames into higher
resolution by predicting high frequency detail. Compared to
image super resolution, video super resolution can
utilize temporal information between frames to get better
result, but also faces an extra challenge to aggregating
information among misaligned frames.
Early methods \cite{Caballero_2017_CVPR, Tao_2017_ICCV} use
optical flow to guide spatial warping of frames.
TDAN~\cite{Tian_2020_CVPR} adopts deformable convolution (DCN)~\cite{Dai_2017_ICCV, Zhu_2019_CVPR}
to align features between frames, and EDVR~\cite{Wang_2019_CVPR_Workshops}
further designed a DCN pyramid to perform
multi-scale feature alignment. BasicVSR++~\cite{Chan_2022_CVPR}
combined optical flow and DCN to generate
high quality DCN offsets for feature alignment.

\subsection{Video Interpolation}

Video Interpolation aims to generate non-existent pixels
temporary between known pixels, in contrast to
Video Super Resolution which generates pixels spatially between
known pixels, so every output pixel need to be warped from
input frames.
Early methods~\cite{Niklaus_2017_CVPR, Niklaus_2017_ICCV}
tried to use an adaptive kernel to perform warping, but is very
inefficient in term of inference speed.
Many flow-based methods~\cite{Jiang_2018_CVPR, Xue2019} use
optical flow to guide warping, but the inaccuracy of predicted
optical flow usually causes distortion to the result, so
extra measures~\cite{Bao_2019_CVPR, 8840983} are usually taken to
refine the warped frame.
There are also methods that generate intermediate frames directly,
like CAIN~\cite{Choi_Kim_Han_Xu_Lee_2020} using channel attention and RIFE~\cite{huang2022rife} using distillation to get good result one simple architecture.

\subsection{Spatio-Temporal Video Super Resolution}

Based on the similar nature of video super resolution
and video interpolation, Spatio-Temporal Video Super Resolution
tries to perform spatial and temporal (interpolation)
super resolution in one network to reach both better result
and better efficiency.
STARnet~\cite{Haris_2020_CVPR} and CycMuNet~\cite{Hu_2022_CVPR}
jointly learns both spatial and temporal SR task using
a mutual learning strategy.
Zooming Slow-Mo~\cite{Xiang_2020_CVPR} proposed deformable ConvLSTM
to explore global temporal contexts. TMNet~\cite{Xu_2021_CVPR}
further proposed a local temporal feature comparison module
to extract short-term motion cues.
YOGO~\cite{10.1145/3503161.3547874} further compact ST-VSR by only
perform alignment once on all input frames to reach
better efficiency.

\subsection{Open Source Neural Network Video Enhancement Tool}

Several open-source tools have been developed to perform video
enhancement tasks that are able to read and write video files.
MMagic\footnote{https://github.com/open-mmlab/mmagic} implemented
video enhancement support using various video super-resolution
and video interpolation models.
Practical-RIFE\footnote{https://github.com/hzwer/Practical-RIFE}
implemented video interpolation using RIFE~\cite{huang2022rife}.
GMFSS\_Fortuna\footnote{https://github.com/98mxr/GMFSS\_Fortuna}
implemented video interpolation based on GMFlow~\cite{xu2022gmflow}.
But these tools simply use OpenCV with its default settings, so
they are only suitable for simple demo purposes.
Anime4K\footnote{https://github.com/bloc97/Anime4K} implemented
simple CNN-based video super resolution and denoising networks
using a shader to achieve real-time enhancement during playback,
but this approach is not suitable for video processing, and
doesn't scale well for more complicated networks.

Under such conditions, VapourSynth gained much attention
thanks to its ability to write video process scripts using Python,
and many VapourSynth scripts and plugins are developed to
perform video enhancement using neural networks.
VapourSynth-Waifu2x-caffe\footnote{https://github.com/HomeOfVapourSynthEvolution/VapourSynth-Waifu2x-caffe}
plugin integrated the Caffe implementation\footnote{https://github.com/lltcggie/waifu2x-caffe}
of the famous waifu2x\footnote{https://github.com/nagadomi/waifu2x}
image super resolution tool.
VSGAN\footnote{https://github.com/rlaphoenix/VSGAN},
vsrife\footnote{https://github.com/HolyWu/vs-rife} and
vs-gmfss\_fortuna\footnote{https://github.com/HolyWu/vs-gmfss\_fortuna}
each implemented VapourSynth processing script using the
PyTorch implementation of
image super resolution model ESRGAN~\cite{wang2018esrgan},
video interpolation model RIFE~\cite{huang2022rife} and
video interpolation tool GMFSS\_Fortuna,
and supports TensorRT speed up using
Torch-TensorRT\footnote{https://github.com/pytorch/TensorRT}.
vs-mlrt\footnote{https://github.com/AmusementClub/vs-mlrt}
plugin further integrates multiple neural network inference engines
into VapourSynth, with support for arbitrary image enhancement
models and RIFE interpolation model.
VSGAN-tensorrt-docker\footnote{https://github.com/styler00dollar/VSGAN-tensorrt-docker}
implemented VapourSynth processing script for various
video enhancement network, and packed into a docker image
for easy use.

\section{Design}

NNVISR provides simple interface for both end user to processing
video, and implementors to provide their network definition.
On top of that, NNVISR also integrates well into current
VapourSynth ecosystem, and can cooperate with other plugins to
build an advanced processing pipeline.

\subsection{Network Integration}

NNVISR designed a simple yet expressive interface for neural
network model to implement, in order to be utilized during
video processing.

NNVISR focus on supporting the kind of video enhancement neural
networks that explores temporal relationship between a definite
number of consecutive frames that we call as frame group.
We separate the whole process into two stages:
the extract stage and the fusion stage.
The extract stage operates on single frame and produces a pyramid
of features in different spatial dimensions (Networks that do not
use feature pyramid can be treated as a pyramid of only one layer).
The fusion stage mixes the features of all frames in a frame group
to produce output frames.

For convenience, we define $n$ to be the number of frames
being ``consumed'' by the network for each inference run. Note that
for interpolation networks, the last frame of a frame group is
usually also the first frame of the next frame group, so in this case
$n$ is the number of input frames minus 1. Based on this definition,
we introduced several flags to precisely describe network behavior:

\begin{itemize}
  \item ``Interpolation'' flag means the network is doing video
  interpolation, and instruct NNVISR to produce output video in
  doubled framerate.
  \item ``Extra frame'' flag means the network accepts $n + 1$
  frames rather than $n$.
  \item ``Double frame'' flag means the network produces $2 n$
  frames for each inference run, otherwise $n$.
  If ``Double frame'' flag is not set and ``Interpolation'' flag is set,
  then input frames are also used as output frames as is,
  and network outputs are all treated as intermediate frames.
\end{itemize}

To integrate neural network into NNVISR, researcher should provide
their network definition in ONNX\footnote{https://github.com/onnx/onnx}
format, which is widely supported by deep learning frameworks.
Notably, PyTorch has built-in support for automatically exporting
model into ONNX format. Specially, we recognized broad usage
of Deformable Convolution~\cite{Dai_2017_ICCV}, especially the V2~\cite{Zhu_2019_CVPR}
variant, among video enhancement networks, mainly due to it's
ability to efficiently adapt to arbitrary object motion using its
offset input, which is crucial in aligning object among frames
and warping frames based on optical flow information.
However, while Deformable Convolution plays an important role
in video enhancement networks, it does not have wide support,
especially is not natively supported by TensorRT. TensorRT has a plugin interface to support custom network operators, so we
implemented the V2 variant of Deformable Convolution for TensorRT.
It is also compatible with PyTorch's Deformable Convolution
implementation provided in torchvision, so networks trained using
PyTorch can be directly used.

NNVISR supports input and output in both RGB and YUV pixel format,
and implementors can choose to support either or both.
RGB is the most common pixel format used in computer vision, while
the YUV420 variant of YUV pixel format is the dominant format for
video storage and processing. Output in YUV pixel format has two
advantages: one being more efficient since YUV420 halves the
spatial dimension of 2 in 3 of its channels, therefore has only
half the amount of data compared to RGB;
another being higher quality because convential conversion
from RGB to YUV420 has been shown to be suboptimal, especially for
high dynamic range contents~\cite{strom2016luma}. By training to
output frame in YUV420 format neural networks can easily achieve
better visual quality compared to conventional conversion method.

NNVISR detects colorspace information in videos and loads the
network trained for the corresponding colorspace, since same
pixel value can and usually mean different colors under
different colorspace. Therefore implementors should train their
networks on datasets in different colorspace separately, so that
the network can produce better result.
Color preciseness is crucial to video processing,
especially on high dynamic range contents, which are gradually
gaining popularity recently.

\subsection{End User Usage}

The usage of NNVISR is simple. NNVISR follows the idiomatic
design of VapourSynth plugin, which provides a single function that
accepts an input video (conventionally called ``clip'' in VapourSynth)
and parameters, and returns the enhanced video.

By setting scale factor and flag parameters to different values, NNVISR
can be configured to perform various kinds of video enhancing tasks:

\begin{itemize}
  \item Setting scale factor to 1 to perform video denoising or .
  \item Setting scale factor > 1 to perform video super resolution.
  \item Specially, setting horizontal scale factor to 1 and
   vertical scale factor to 2 to perform video deinterlacing.
  \item Setting scale factor to 1 and ``Interpolation'' flag to perform
   video interpolation.
  \item Setting scale factor to 1, ``Interpolation'' flag and
   ``Double frame'' flag to perform video interpolation along with
   denoising.
  \item Setting scale factor > 1, ``Interpolation'' flag and
  ``Double frame'' flag to perform spatial-temporal video resolution.
\end{itemize}

Usually real-world videos contains more than one scene while
video enhancement models are only trained on a single scene.
Therefore when crossing the scene boundary, enhancement result is
usually bad so this should be avoid. VapourSynth framework has a
standard way to signal the occurence of scene change, therefore
user can choose the most suitable plugin to detect scene change
based on the video content and user's need. There are already
multiple filters that detects scene change using traditional
method, and employing a neural network to detect scene change
can also be easily implemented.

In addition to source code, we also publish pre-built binaries for
Windows and Linux as conda packages, so that users can easily install
NNVISR without going through the complicated environment configuations.

\subsection{Internal Design}

NNVISR internally performs the grouping of frames, taking account of
$n$ of network and scene change. For interpolation networks, NNVISR
uses a carefully designed frame layout to reduce unnecessary memory
usage as much as possible.

NNVISR always grouping frames within a scene, and has a complete set
of rules to always output required number of frames.
For interpolation network, NNVISR duplicates the last frame of the scene.
If the last group doesn't have enough frames for network, last frames
from the previous group is recycled to provide enough input frames.
If the whole scene does not contain enough input frames then the last
frame is further duplicated to satisfy network input count.

\begin{figure}

\begin{center}
\begin{tabular}{c|ccc|ccc|c}
  & 3 & 1 & 2 & 9  & 7  & 8  & ... \\
0 & 6 & 4 & 5 & 12 & 10 & 11 & ...
\end{tabular}
\end{center}
\caption{NNVISR frame store layout where $n = 3$ and $b = 2$.
Numbers denote the frame index.
Actual store order is column first.} \label{fig:layout}
\end{figure}

Interpolation networks usually require $n + 1$ input frames, and for a
batch of $b$, the naive way need to provide $b * (n + 1)$ frames to the
network. However for a batch of $b$, the total number of related frames
is actually $b * n + 1$, and $b - 1$ frames are duplicated and taking
extra space. The nature of batching requires the same input from different
batch placed at contiguous location in memory. Rather than placing the
frames in the natural order, NNVISR use an layout shown in Fig. \ref{fig:layout}.
This layout eliminates the need to duplicate frames and can be extended
to any $b$. To process the next batch, only the last frame need to be
copied to the location of the second last frame (6 to 5 in the figure).

\section{Conclusions}

We have publicly released NNVISR, a versatile
video processing tool that supports arbitrary networks and
various kinds of video enhancing tasks.
Moreover, it's high performance, well documented, easy to use
and integrates well into current video processing toolchain.
In this report, we introduced the design of NNVISR, as well as
the considerations behind. We hope NNVISR can help the
adoption of video enhancement research echievements into
downstream industrial applications.


\bibliographystyle{ACM-Reference-Format}
\bibliography{main}

\end{document}